\documentclass[journal,web]{IEEEtran}
\usepackage{url,lineno,microtype,subcaption}
\usepackage{times}
\usepackage{epsfig}
\usepackage{graphicx}
\usepackage{amsmath}
\usepackage{amssymb}
\usepackage{algorithm}
\usepackage{algorithmicx}
\usepackage{algpseudocode}
\usepackage{bm}
\usepackage{color}
\usepackage{multirow}
\usepackage{makecell}
\usepackage{url}
\usepackage[colorlinks,linkcolor=blue]{hyperref}

\title{ES-ImageNet: A Million Event-Stream Classification Dataset for Spiking Neural Networks}

\author{Yihan Lin\,$^{1}$, Wei Ding\,$^{1}$ Shaohua Qiang\,$^{1}$ Lei Deng\,$^{1}$ Guoqi Li$^{1,*}$
\thanks{
All the authors are with Center for
 Brain-Inspired Computing Research, Beijing,
 Department of Precision Instrument, Tsinghua  University, P. R. China. 
 The Corresponding authors: Guoqi~Li. (E-mail: liguoqi@mail.tsinghua.edu.cn).}%
}

\begin{document}
\maketitle
\markboth{Endorsed publication: Frontiers in NeuronScience}  
{Lin \MakeLowercase{\textit{et al.}}: ES-ImageNet: A Million Event-Stream Classification Dataset for Spiking Neural Networks}
\begin{abstract}
With event-driven algorithms, especially the spiking neural networks (SNNs), achieving continuous improvement in neuromorphic vision processing, a more challenging event-stream-dataset is urgently needed. However, it is well known that creating an ES-dataset is a time-consuming and costly task with neuromorphic cameras like dynamic vision sensors (DVS). In this work, we propose a fast and effective algorithm termed Omnidirectional Discrete Gradient (ODG) to convert the popular computer vision dataset ILSVRC2012 into its event-stream (ES) version, generating about 1,300,000 frame-based images into ES-samples in 1000 categories. In this way, we propose an ES-dataset called ES-ImageNet, which is dozens of times larger than other neuromorphic classification datasets at present and completely generated by the software. The ODG algorithm implements an image motion to generate local value changes with discrete gradient information in different directions, providing a low-cost and high-speed way for converting frame-based images into event streams, along with Edge-Integral to reconstruct the high-quality images from event streams. Furthermore, we analyze the statistics of the ES-ImageNet in multiple ways, and a performance benchmark of the dataset is also provided using both famous deep neural network algorithms and spiking neural network algorithms. We believe that this work shall provide a new large-scale benchmark dataset for SNNs and neuromorphic vision. 
\end{abstract}
\section{Introduction}

In recent years, Spiking Neural Networks (SNNs) have attracted extensive attention in the fields of computational neuroscience, artificial intelligence and brain-inspired computing  \cite{TSMINC,TIANJI}. As called the third generation neural network  \cite{SpikingNeuron}, SNNs have the ability to process spatiotemporal information and own stronger biological interpretability than Artificial Neural Networks (ANNs, or Deep Neural Networks). They have been applied in quantities of tasks, such as pattern recognition  \cite{Speech_recog_LSM,Speech_recog_LSM2,rouat2013spatio}, high-speed object tracking  \cite{DashNet}, optical flow estimation  \cite{paredes2019unsupervised} with the help of neuromorphic hardware such as TrueNorth  \cite{TURENORTH}, Loihi  \cite{LOIHI}, DaDianNao  \cite{DaDianNao}, Tianjic  \cite{TIANJI}. In recent years, the continuous expansion of datasets in image classification  \cite{Imagenet,cifar,MNIST}, natural language processing  \cite{MSMARCO,Squad} and other fields has been challenging the ability of AI and promoting the development of AI. The researchers hope that AI can surpass humans in the corresponding tasks. However, for the SNNs, the research is still in the rising stage with obstacles gradually appearing, where the lack of suitable datasets is one of the biggest ones. We now have \emph{N-MNIST}  \cite{Cal101}, \emph{N-Caltech101}  \cite{Cal101}, \emph{DVS-Gesture}  \cite{gesture}, \emph{CIFAR10-DVS}  \cite{dvscifar10} and other neuromorphic datasets (or event-stream datasets, ES-datasets), but those existing datasets designed for SNNs are relatively small in scale. As more algorithms are proposed, the scale of SNNs is growing larger. Therefore the existing datasets have been difficult to meet the demand for training and validation of SNNs.

\begin{figure*}[!t]
    \centering
    \includegraphics[width=1 \linewidth]{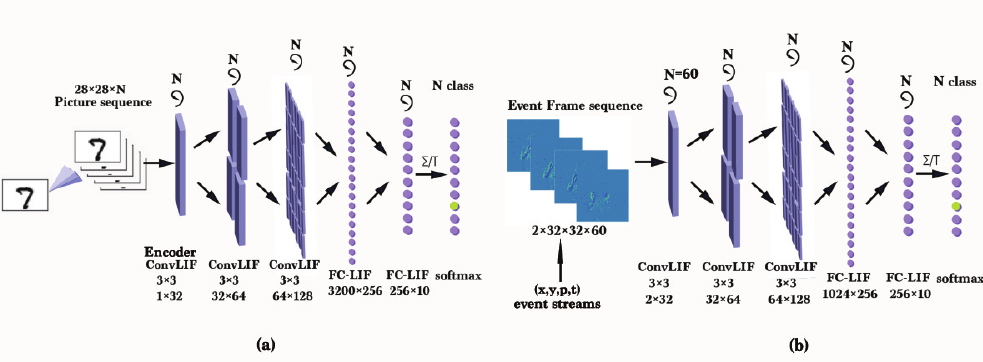}
    \caption{{\bf (a)} {\bf An approach for training the LIF-SNN  \cite{LIF} on an ANN-oriented dataset}. Here the SNN uses the Rate Coding and an ANN-like structure, so it can be trained using frame data naturally. {\bf (b)} {\bf Train an LIF-SNN with GPUs on a DVS-dataset (ES-dataset recorded by DVS)}. Here we need to accumulate events within a small period as an event frame and get an event frame sequence with $N$ frames for training. On the neuromorphic processor, those asynchronous event data can be processed more efficiently.}
    \label{fig:mnist_struct}
\end{figure*}

A compromised solution towards this problem is to train SNNs on the large-scale traditional static datasets directly. Taking image classification for an instance, the common method is copying an image multiple times to form an image sequence, and then the sequence is fed to the spike encoding layer of an SNN, as {\bf Fig. \ref{fig:mnist_struct} (a)} shows. However, there is an obvious shortcoming that the data redundancy makes the training cost increase many times without any effective information being added. For comparison, the way to train an SNN on an ES-dataset is also shown in {\bf Fig. \ref{fig:mnist_struct} (b)}. Compared to the common method, it is more natural for SNNs to process such sparse and temporal data by making full use of temporal characteristics. So the datasets inspired by neuromorphic visual sensor imaging mechanism are still considered to be the most suitable datasets for SNNs. 

\begin{table*}[bt]
 \centering
  \centering
\caption{Comparison between existing ES-datasets and ES-ImageNet}
   \label{tab:Comparison}
\begin{tabular}{m{4cm}<{\centering}m{3cm}<{\centering}m{2cm}<{\centering}m{3cm}<{\centering}m{1.5cm}<{\centering}m{1.5cm}<{\centering}}

\hline
\makecell*[c]{\textbf{Name} } &\makecell*[c]{ \textbf{generating speed}$^{\rm a}$}& \makecell*[c]{\textbf{Resolution} }  & \makecell*[c]{\textbf{\# of samples}} & \makecell*[c]{\textbf{\# of class}} & \makecell*[c]{\textbf{\# Type} }\\ \hline

\rule{0pt}{16pt}  POKER-DVS \cite{6497055}         & -                    & 32$\times$32          & 131                                & 4           & Classify        \\

\rule{0pt}{16pt}  N-MNIST \cite{Cal101}            & 300 ms/sample          & 28$\times$28       & 60, 000 training + 10, 000 test   & 10           & Classify            \\

\rule{0pt}{16pt}  DVS-Caltech101 \cite{Cal101}       & 300 ms/sample    & 302$\times$245 in average  & 8709            & 100                      & Classify       \\

\rule{0pt}{16pt}  {DVS-UCF-50 \cite{ETHRPG}} & {6800 ms/sample}     & {240$\times$180}          & {6,676} & {50}  
& Classify \\ 

\rule{0pt}{16pt}  {DVS-Caltech-256 \cite{ETHRPG}} & {1010 ms/sample}   & {240$\times$180}  & {30,607} & {257}      
& Classify  \\ 

\rule{0pt}{16pt}  {DVS-VOT-2015 \cite{ETHRPG}} & {30 FPS, 20.70 s/sample}   & {240$\times$180}  & {67} & {-}      
& Track  \\ 

\rule{0pt}{16pt}  DVS-CIFAR10 \cite{dvscifar10}          & 300 ms/sample                           & 512$\times$512           & 10,000                   & 10                      & Classify        \\

\rule{0pt}{16pt}  DVS-Gesture \cite{gesture}          &  6 s/sample       & 128$\times$128           & 1,342              & 11                        & Classify      \\

\rule{0pt}{16pt}  {Pred-18 \cite{pred18}} & { 15 FPS}   & {240$\times$180}  & {1.25h (67.5k Frames) } & {2}      
& Detect  \\ 

\rule{0pt}{16pt}  {Action Recognition \cite{miao2019neuromorphic}} & {5 s/sample }   & {346$\times$260}  & {450} & {10}      
& Classify  \\ 

\rule{0pt}{16pt}  {1Mpx Detection Dataset \cite{largeDetect}} & {60 s/sample}   & {304$\times$240}  & {14.65h, 255,781 object} & {2}      
& Detect  \\ 

\rule{0pt}{16pt}  {SL-ANIMALS-DVS \cite{SLANIMALS}} & { - }   & {128$\times$128}  & {1,102} & {10}      
& Classify  \\ 

\rule{0pt}{16pt}  {DVS-Gait-Day/Night \cite{wang2021event}} & {3-4 s/sample  }   & {128$\times$128}  & {4,000} & {20}      
& Classify  \\ 

\rule{0pt}{16pt}  {N-ROD \cite{cannici2021n}} & {6.6 s/sample  }   & {256$\times$256}  & {41,877} & {51}      
& Classify  \\ 

\rule{0pt}{16pt}  \textbf{ES-ImageNet} & \textbf{29.47 ms/sample}$^{\rm b}$     & \textbf{224$\times$224}$^{\rm c}$           & \textbf{1,257,035 training + 49,881 test} & \textbf{1000}                         
& Classify   \\ 
\hline

\end{tabular}
\footnotesize{$^{\rm a}$ The average time taken for generating each sample  or average recording time (for directly recorded).}\\
\footnotesize{$^{\rm b}$ $Threshold = 0.18$.}\\
\footnotesize{$^{\rm c}$ The events are generated in a range of 256$\times$256-pixel. But only those in central 224$\times$224-pixel are meaningful, while others are noise generated by image edge's motion.}
\end{table*}

Since SNNs have benefited from neuromorphic data, efforts are also devoted to recycling the existing RGB-cameras datasets to generate neuromorphic datasets. Mainly there are two different methods for this task. One is to use dynamic vision sensor (DVS) cameras to record a video generated from raw data with an LCD screen \cite{Cal101,dvscifar10}. This method is time-consuming and costly, which is impossible for manufacturing large-scale dataset. The other one is to generate the event data using software to simulate the principle of DVS cameras \cite{PIX2NVS,RVD}. This kind of method is more suitable for generating large-scale event base datasets. However, the data redundancy caused by the existing converting algorithms increases the volume of the datasets. In this work, we optimize the existing algorithms of the second method to obtain the dataset with less redundancy.


In this way, an ES-dataset converted from the famous image classification dataset \emph{ILSVRC2012} \cite{ILSVRC15} is generated, which is named \emph{event-stream ImageNet} or \emph{ES-ImageNet}. In \emph{ES-ImageNet}, there are about 1.3 M samples converted from \emph{ILSVRC2012} in 1000 different categories. {\bf \emph{ES-ImageNet} is now the largest ES-dataset for object classification at present}. We have sorted out the information of representative existing ES-datasets and compared them with \emph{ES-ImageNet}, the results are summarized in {\bf Table \ref{tab:Comparison}}. Owing more categories and samples also make it probably the most challenging classification ES-dataset, providing space for continuous improvement of event-driven algorithms.

A good conversion algorithm is expected to generate a dataset that is not larger than the source. If it is not required to imitate the characteristics of DVS, the optimal binary-coding conversion is able to encode the original information with the same size of data. So when the conversion algorithm generates a larger dataset than the original one, there must be data redundancy. In order to simulate the DVS cameras, we can allow a little redundancy. However, most of the existing conversion methods generate a much larger dataset (for example, N-MNIST \cite{Cal101} and Flash-MNIST's storage volume is tens of GB, where the original MNIST is no larger than 100MB). This is far from the original intention of DVS sparsity, and it is not conducive to high-speed efficient processing and large-scale applications. So a simple bio-inspired algorithm called \emph{Omnidirectional Discrete Gradient} (ODG) is applied. This algorithm captures the sequential features of images and then places them on the time axis with timestamps to generate event streams. It reduces the information redundancy, which brings higher generation speed and less data redundancy than the existing conversion algorithms. It can be regarded as the streamlining of random saccades for deep learning use, where the latter one is a common bio-inspired generation methods.

To guarantee a suitable sparsity of data and the amount of information, we also conduct preparatory experiments to control the event rates and the amount of information of the generated samples. Further analysis about the computation cost of different algorithms is conducted, which supports the dataset to be an SNN-friendly dataset.

The main contributions of this work are three-fold. 

{\bf (i)} We introduce a new large-scale ES-dataset named \emph{ES-ImageNet}, which is aimed at examining SNNs' ability to extract features from sparse event data and boosting research on neuromorphic vision. This work shall provide a new large-scale benchmark dataset for SNNs and neuromorphic vision tasks.

{\bf (ii)} A new algorithm called ODG is proposed for converting data to its event stream version. We consider the guiding ideology behind it to be a paradigm for conversion from static data to ES-data, which avoids data redundancy.

{\bf (iii)} Several ways for analyzing the dataset are provided, including  information loss analysis using 2D information entropy (2D-Entropy) and the visual perception of the reconstructed pictures. Also two preparatory experiments are designed for designing the algorithm, which may provide inspiration for further improvement.

\subsection*{Related work}

{\bf ES-datasets collected directly from the real scenarios}. DVS cameras can generate unlabeled ES data directly  \cite{dvscamera}. The ES data is often organized as a quad $(x, y, t, p)$, where $(x, y)$ is the topological coordinates of the pixel, $t$ is the time of spike generation, $p$ is the polarity of the spike. Such datasets are easy to generate and close to practical application scenarios, like datasets that can be used for tracking and detection  \cite{pred18,simultaneous,largeDetect}, datasets for 3D scene reconstruction  \cite{carneiro2013event,kim2016real}, neural morphology datasets for optical flow estimation  \cite{simultaneous,EVF}, and datasets for gesture recognition  \cite{gesture}. Due to the high sampling rate and authenticity, this kind of datasets are of great help to the development of the applications in the high speed scenes. But because of the huge workload of making the real scenario recording datasets, their sizes are often small, which is difficult to meet the demand of examining deep SNNs algorithm.
 
{\bf Transformed ES-datasets with help of neuromorphic sensors}.  These datasets are mainly generated by the labeled static image datasets through neuromorphic sensors. Different from the first ones, this kind of datasets are mainly obtained from the datasets which have been widely studied and used for traditional ANN tasks, such as \emph{N-MNIST}  \cite{Cal101}, \emph{DVS-UCF-50}, \emph{DVS-Caltech-256}  \cite{ETHRPG}, and \emph{CIFAR10-DVS}  \cite{dvscifar10}. In order to make such datasets, one way is to use a screen to display a static picture, then face the DVS camera to the screen and move the camera along the designed trajectory to generate events. Because of the similarity between the transformed dataset and the original one, this kind of dataset can be used and evaluated easily. Therefore, they are also the most widely used datasets in SNN research. However, while in the recording process, noise is introduced, especially that caused by flashing LCD screen.

{\bf Completely software-generated ES-datasets without neuromorphic sensors}. The algorithms are used to simulate the characteristics of DVS cameras with labeled data here. The dynamic sensors can capture the dynamic information from the video streams or picture sequences, while this process can also be completed by specific algorithms \cite{PIX2NVS,RVD,DashNet}. These methods can avoid a large number of experiments needed for collecting data. However, the existing algorithms used for converting static-data to event-data always extract information with too much redundancy that is brought by the randomness or repetitiveness of the generation algorithms. 
\section{Materials and Methods}

In this section, we will introduce a method to generate the \textit{ES-ImageNet} with a corresponding reconstruction method, including the color space conversion, ODG processing, hyper-parameters choosing, and sparse storage.\\
\quad
\subsection*{Color Space Conversion}

Traditional ES-datasets utilize DVS cameras to record the changes of intensity asynchronously in ES format, which encode per-pixel brightness changes. In RGB (Red-Green-Blue) color models, a pixel's color can be described as a Triplet $(Red, Green, Blue)$ or $(R, G, B)$, which does not indicate brightness information directly. When using HSV (Hue-Saturation-Value) color model, it’s described as $(Hue, Saturation, Value)$ or $(H, S, V)$. Generally, the images in \emph{ILSVRC2012} dataset are stored in RGB color space, therefore images need to be converted to HSV color space, as shown in 

\begin{equation} 
\label{equa:csv_start}
H=\left\{
\begin{aligned}
0^{\circ} & ~ & if~~M=m\\
60^\times \frac{G-B}{M-m} + 0^{\circ} & ~ & if~~ M = R ~ and~  G > B\\
60^{\circ}\times \frac{G-B}{M-m} + 360^{\circ} & ~ & if~~ M = R ~ and~  G  \leq B\\
60^{\circ}\times \frac{G-B}{M-m} + 120^{\circ} & ~ & if~~ M = G\\
60^{\circ}\times \frac{G-B}{M-m} + 240^{\circ} & ~ & if ~~M = B
\end{aligned}
\right.
\end{equation}

\begin{equation} 
S=\left\{
\begin{aligned}
0  & ~ & if~~\max=0\\
\frac{M-m}{M} = 1 -  \frac{m}{M} & ~ & otherwise\\
\end{aligned}
\right.
\end{equation}

\begin{equation} 
V=M,
\label{equa:csv_end}
\end{equation}
where $M = \max\{R, G, B\}$ and $m = \min\{R, G, B\}$. In this algorithm, we use $V$ as a reference of light intensity. In the HSV hex-cone model, the value indicates the brightness of the color. And for the light source, the value is also related to the brightness of the illuminant, so it can be used as a reference for light intensity.

\quad
\subsection*{Event Generator}
\quad

\quad 
\begin{algorithm}
\caption{ODG event generator}
\label{alg:onlyone}
\begin{algorithmic}
	\Require $Image$ 
	\Ensure $Stream$
	\State xTrace = [1,0,2,1,0,2,1,1,2],   yTrace = [0,2,1,0,1,2,0,1,1],   Thresh = 0.18,   T=8
	\Function {Generator}{Image}
		\State W = Image.size[0], H = Image.size[1]
		\State Image =  zeroPadding(upSampling( Image, (254, 254) ) , 2)
		\State V = RGB2HSV(Image).V
                	\For {t = 0 $\to$ T}
			\State x = xTrace[t], y = yTrace[t]
			\State newImage = V[x : x+W, y : y+H]
			\If {t $>$ 0}
				\State oldX = xTrace[t-1], oldY = yTrace[t-1]
				\State ImgDiff = newImage - lastImage
              			\State posEvent = ImgDiff(ImgDiff $\geq$ Thresh),  negEvent = ImgDiff(ImgDiff $\leq$ -Thresh)
				\For{ i = 0 $\to$ len(posEvent)}
					\State Ex = posEvent[0], Ey = posEvent[1]
					\If {(Ex, Ey) is in valid range}
						\State posStream.append( (Ex, Ey,t) )
					\EndIf
				\EndFor
				\For{ i = 0 $\to$ len(negEvent)}
					\State Ex = negEvent[0], Ey = negEvent[1]
					\If {(Ex,Ey) is in valid range}
						\State negStream.append( (Ex,Ey,t) )
					\EndIf		
				\EndFor
			\EndIf
                         	\State lastImage  = newImage
                \EndFor
	\EndFunction
\end{algorithmic}
\end{algorithm}

To stimulate the intensity changes, we use ODG here. Based on the truth that  animals like toads or frogs can only respond to moving objects \cite{ewert1974neural}, we believe that we can obtain the necessary information for object recognition by imitating the frog retinal nerves, specifically, ganglion cells that generate features. Three important kinds of ganglion cells act as edge detectors, convex edge detectors, and contrast detectors, generating sparse local edge information. This inspires the main idea of ODG, which is artificially changing the light intensity and detecting the necessary local edge information in multiple directions.

\begin{figure*}[!t]
    \centering
    \includegraphics[width=1  \linewidth]{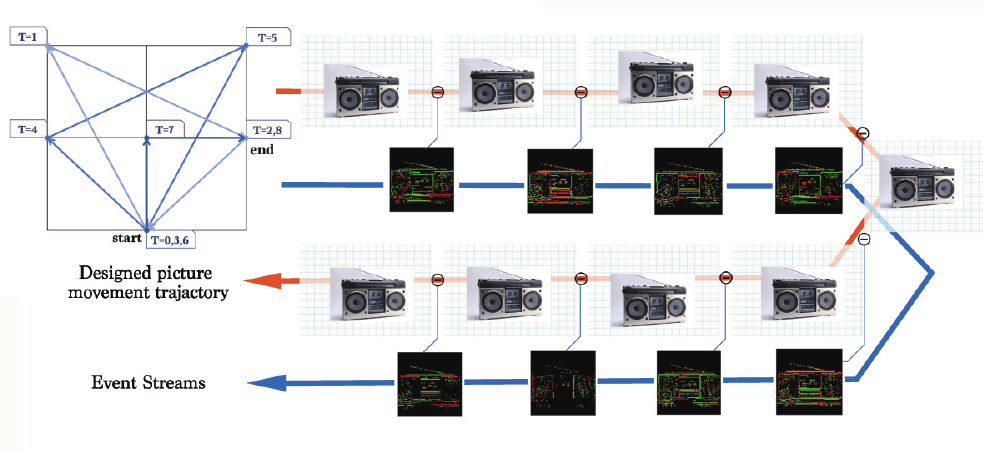}
    \caption{{\bf The moving trajectory of images used to generate the events}. The numbers in the small blue squares is the timestamp when an image reaches the corresponding position. The pipeline shows a complete process of generating event stream.}
    \label{fig:algorithm_intro}
\end{figure*}

Different from random saccades generation used widely \cite{ETHRPG}, we only choose necessary directions in a fixed order, and the necessary number of frames to minimize the data redundancy and we will explain it later. This algorithm generates an event stream for each picture in \emph{ILSVRC2012} with a specific moving path shown in {\bf Fig. \ref{fig:algorithm_intro}}, and the algorithm is summarized in {\bf Algorithm \ref{alg:onlyone}}. The trigger condition of the events is described in 

\begin{equation} 
p(x,y,t)=\left\{
\begin{aligned}
-1  ~~~~~ &if~~V(x,y,t)-V(x,y,t-1)<-Thresh\\
~1  ~~~~~ &if~V(x,y,t)-V(x,y,t-1)>Thresh,\\
\end{aligned}
\right.
\label{equa:logic}
\end{equation}
where $p(x,y,t)$ denotes the polarity of the event at $(x,y,t)$, $V$ is the value of pixel and $Thresh$ is the difference threshold. This algorithm only involves linear operations with time complexity of $\mathcal{O}(W^2T)$, where $W$ denotes the width of the image and $T$ is the length of time. The ES-ImageNet is generated without randomness so that users can reconstruct the original information using the path information and design data augmentation freely.

In {\bf Algorithm \ref{alg:onlyone}}, there are four hyper-parameters to be selected: a sequence of x coordinate ($xTrace$), a sequence of y coordinate ($yTrace$), the difference threshold ($Thresh$) in {\bf Eq. (\ref{equa:logic})} and the number of time steps ($T$). We designed two preparatory experiments to determine these hyper-parameters.

\quad
\quad
\subsection*{Select the Hyper-parameters}
\quad

\begin{figure*}[!t]
    \centering
    \includegraphics[width=1.0  \linewidth]{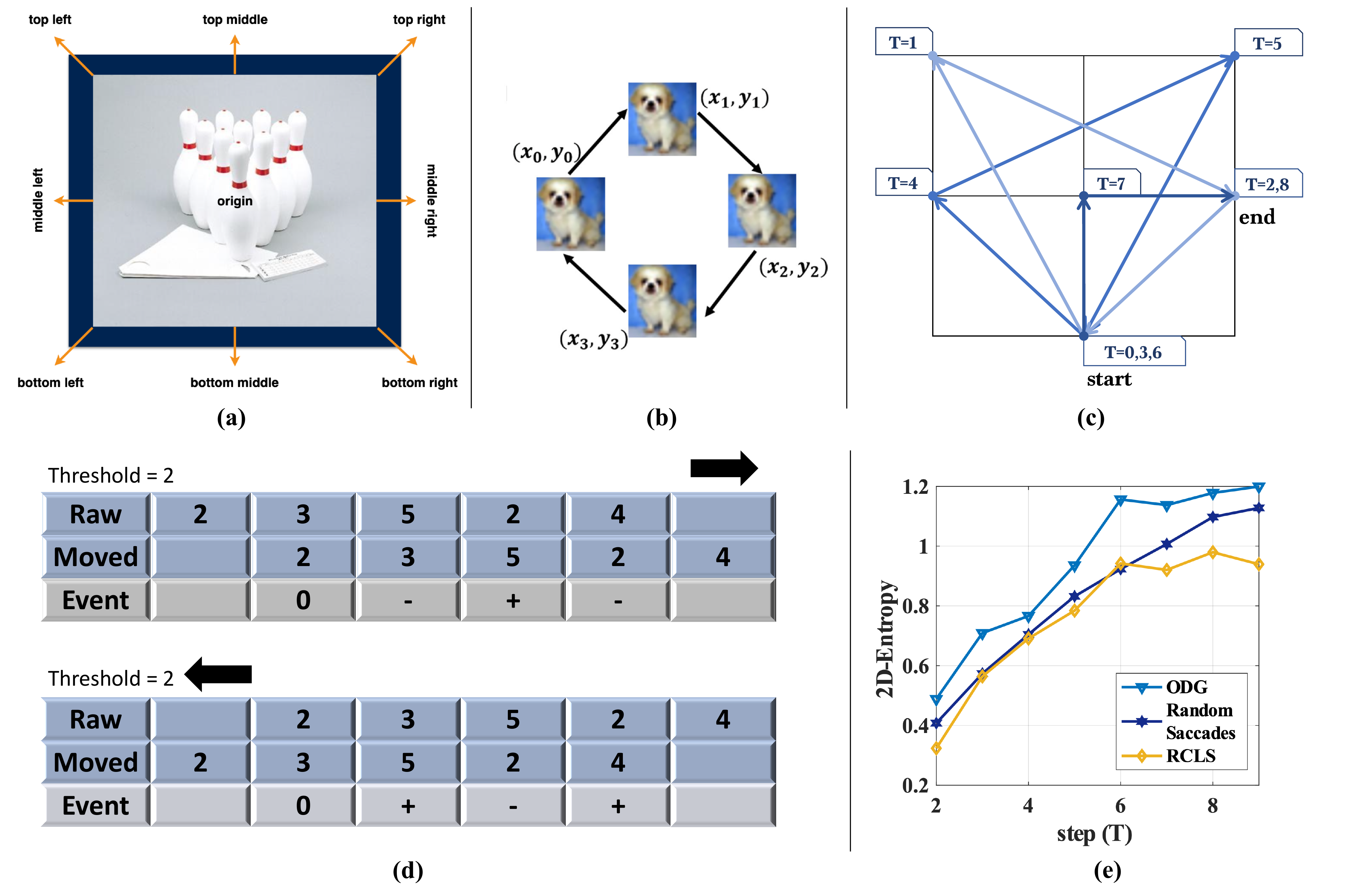}
    \caption{{\bf Comparison of three different image motion}. (a) The candidate moving directions used in random saccades generation of DVS-Caltech-256 and DVS-UCF-50 \cite{ETHRPG}. (b) The path used in RCLS of DVS-CIFAR-10 \cite{dvscifar10}. (c) Trajectory used in the ES-ImageNet. (d) Illustration for that the opposite directions in the generation path would only generate opposite events. (e) 2D-Entropy comparison among the three generating paths with different steps ($T$). ODG is superior to the other two methods in the sense of 2D-Entropy based on the reconstructed gray images.}
    \label{fig:algorithm}
\end{figure*}

{\bf Trajectory.} The choice of the path is important, which includes designing $xTrace$ and $yTrace$ along with choosing $T$. Most of the existing conversion methods choose fast random saccades or repeated fixed paths. The former selects eight directions for simulating fast eyes movement (random saccades), while the latter uses only four directions (repeated closed-loop smooth (RCLS)), as shown in {\bf Fig. \ref{fig:algorithm} (a)-(c)}.

To verify the information obtained by these different methods, we evaluate it by comparing their 2D information entropy (2D-Entropy), which is positively correlated with the amount of information in an image. The average neighborhood gray value of the image is selected to represent the spatial characteristics, and a $3\times 3$ field is commonly used. The feature pair $(i, j)$ is used to calculate 2D-entropy, where $i$ is the gray value of the pixel, and $j$ is the rounding down of the mean neighborhood gray value. The 2D-Entropy of gray image is defined as {\bf Eq. (\ref{equa:2dH})}, where $p_{(i,j)}$ denotes the frequency of the feature pair $(i, j)$ and $g$ is the gray level.

\begin{equation} 
H= \sum_{i=0}^{g}\sum_{j=0}^{g}-p_{(i,j)}log_2(p_{(i,j)}).
\label{equa:2dH}
\end{equation}

Because these algorithms use the frame difference for event generation, and the adjacent frames are actually the same image, the movement with the opposite direction would always generate events with opposite polarity. Therefore, a new step with an opposite direction to the existing movement does not add more effective information into the sample, and that is what the existing algorithms can be optimized. As shown in {\bf Fig. \ref{fig:algorithm} (d)}, the number in the cell denotes the pixel value. A row of pixels move left or right and calculate the difference under the same threshold would only obtain a series of events with exactly the opposite polarity.

Based on this observation, we avoid the repeated or the opposite path design in the \emph{ODG}. Furthermore, to quantitatively illustrate the benefits, we randomly select 100 images from ImageNet-1K, extract events in different $T$ with the three different methods, and then reconstruct them into gray images to calculate 2D-Entropy. In this way we get {\bf Fig. \ref{fig:algorithm} (e)}, and the higher curve of ODG may support our modification. 

Through analyzing the information (2D-Entropy) curves calculated for each method over several time steps in {\bf Fig. \ref{fig:algorithm} (e)}, we find that the 2D-Entropy increases slowly after $T \geq 6$, but the size of the dataset would still increases linearly with $T$ getting larger. In order to make a balance between the temporal characteristics, the amount of information and the size of dataset, we set $T=8$.

\begin{figure}[!t]
    \centering
    \includegraphics[width=0.6  \linewidth]{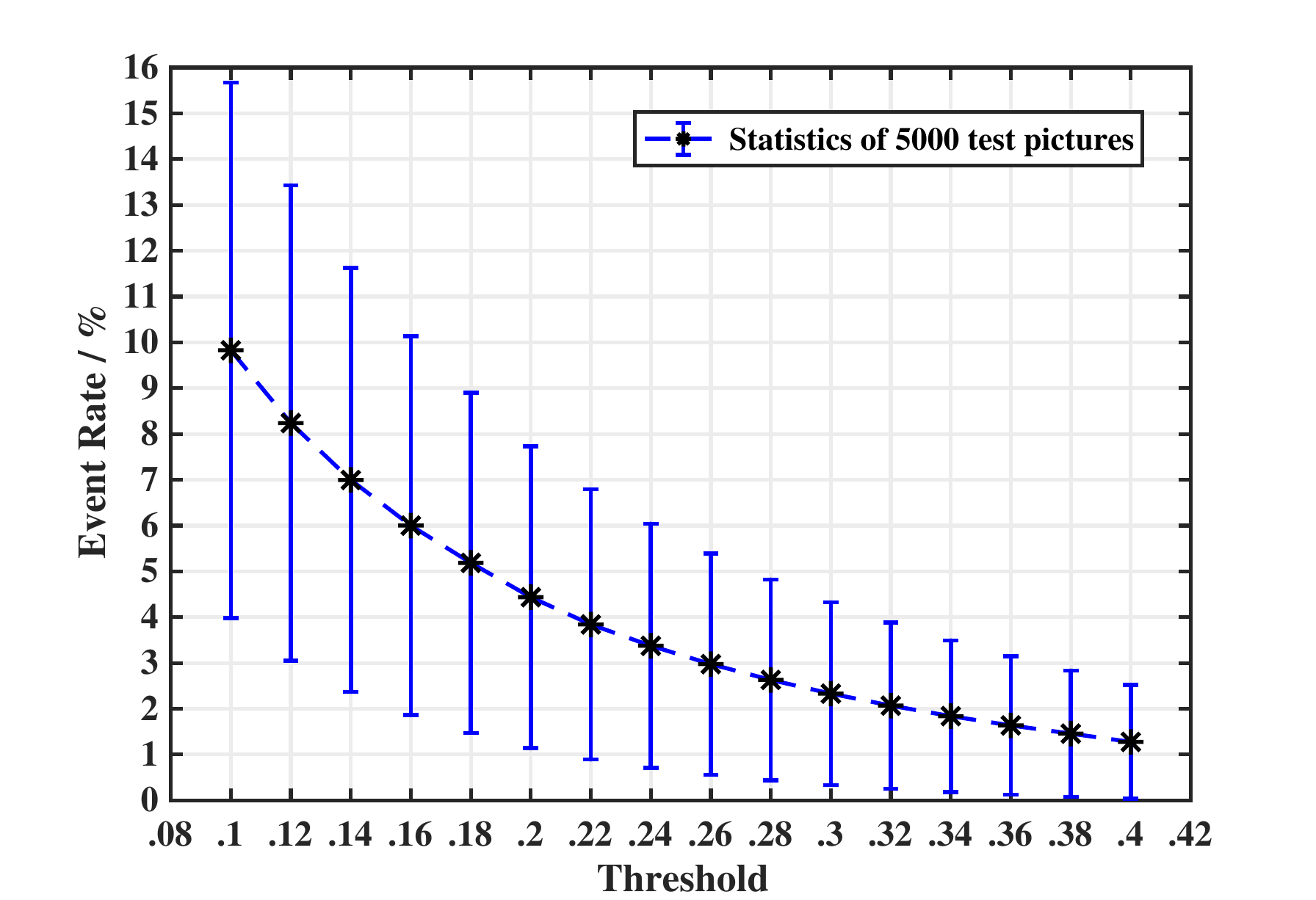}
    \caption{{\bf A preparatory experiment for determining the threshold in ODG }. We observed that the event rate shows a trend of an exponential decrease in the threshold. In consideration of the event rate of generated samples and to avoid generating too many invalid samples that have extremely low event rate, we choose $threshold=0.18$.}
    \label{fig:select}
\end{figure}

{\bf Threshold.} An important indicator for ES-dataset is event rate, which is defined as the proportion of pixels that have triggered event. The most influential parameter for event rate is the threshold $Thresh$ (when the motion path is fixed). Because of the high correlation of brightness between adjacent pixels, it is hard to estimate the distribution of the difference between adjacent pixels using the static method, so a preparatory experiment is needed. We randomly select 5 pictures from each category and get 5000 pictures. The threshold in the experiment varies from 0.1 $\sim$ 0.4. The results are shown in {\bf Fig. \ref{fig:select}}. After many tests, we choose $0.18$ as the threshold value, for an estimated event rate of $5.186\%$, with the event rate of most samples being in the range of $1\%$ to $10\%$. This result will be verified on the whole dataset. It should be noted that many events may be generated by the movement of the edge of the image, and they have been wiped out. 

In addition, the original images' longest sides are normalized to 255 by the nearest-interpolation algorithm. The nearest interpolation is a mapping from the normalized coordinates after zooming to the integer grid coordinates. The generated event stream version training set is 99GB and test set is 4.2GB, which are stored in quad format $(x,y,t,p)$. If converted to the frame version like the short video, the size of the whole dataset can be further reduced to 37.4GB without information loss. For ease of use, we store all of these tensors as a file in .npz format, using the scientific computing package "Numpy" of Python. The event-frame format version is more suitable for deep learning, and we will also provide this version, while the quad format version is the classical ES-dataset.\\

\subsection*{Data Analysis}

{\bf Event Rate.} To examine the quality of the data, we calculate the event rates of the whole generated dataset and summarize them in {\bf Table \ref{table:eventrate}}. It can be seen that the pixels which trigger the events are only about one-twentieth of all pixels. And from this point of view, the prediction obtained from the preparatory experiment is accurate. Since our events are generated from the images processed by nearest-neighbor interpolation, our event rate statistics are also calculated in this range. When we use the training data, we often place the positive events and negative events in different channels and organize them in the unified $2\times224\times224~(C\times W \times H)$ frame format. Therefore,  we re-calculate the event rate during the  training process in the {\bf Table \ref{table:eventrate}}, which is lower than that of generating process and be more meaningful for training process. 

\begin{figure*}[bt]
    \centering
    \includegraphics[width=0.6  \linewidth]{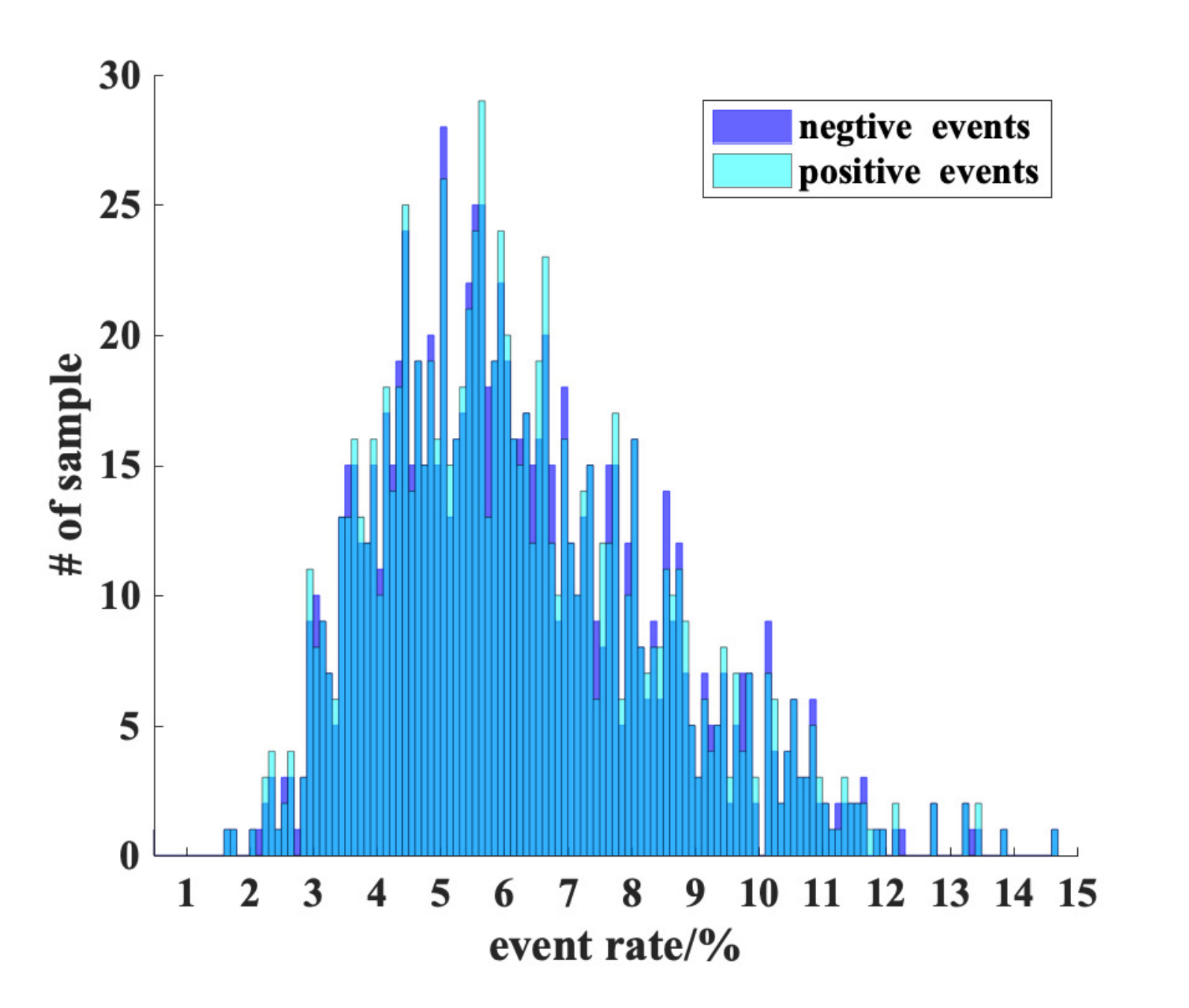}
    \caption{{\bf A detailed inspection about fire rate}. Most samples have 5\% to 6\% event rate, and this figure shows a significant skew distribution. A sample with a 5\% event rate is also a relatively sparse level when recorded by event cameras.}
    \label{fig:Distribute2}
\end{figure*}

Furthermore, we calculate the distribution histogram of positive and negative events and show it in {\bf Fig. \ref{fig:Distribute2}}. The results in the figure show that the distribution of positive and negative events is very close, which may be because most of the entities in the original images are represented as closed graphics.

\begin{table}[hbt]
\centering
\caption{Event rate of ES-ImageNet.}
\centering
\renewcommand\arraystretch{1.5}
\begin{tabular}{lclclclclcl}  
\hline
\textbf{Generating Process}                   & \multicolumn{2}{l}{\textbf{Training set}} & \multicolumn{2}{l}{\textbf{Testing set}} \\ \hline
\multicolumn{1}{l|}{}       & Mean               & $\sigma$             & Mean   
            & $\sigma$            \\\hline
\multicolumn{1}{l|}{Events}  & 5.215\%         & 3.776\%        & 5.385\%        & 3.837\%        \\ 
\multicolumn{1}{l|}{ON}     & 5.211\%          & 3.777\%        & 5.385\%         & 3.838\%        \\ 
\multicolumn{1}{l|}{OFF}   & 5.22\%          & 3.78\%        & 5.38\%         & 3.84\%        \\ 
\hline
\textbf{Event-frame Format}                   & \multicolumn{2}{l}{\textbf{Training set}} & \multicolumn{2}{l}{\textbf{Testing set}} \\ \hline
\multicolumn{1}{l|}{}       & Mean               & $\sigma$             & Mean               & $\sigma$            \\\hline
\multicolumn{1}{l|}{Events} & 4.461\%             & 3.560\%               & 5.231\%             & 3.770\%             \\
\multicolumn{1}{l|}{ON}     & 4.458\%             & 3.570\%               & 5.229\%             & 3.770\%              \\
\multicolumn{1}{l|}{OFF}    & 4.460\%             & 3.560\%               & 5.230\%             & 3.770\%              \\  
\hline
 \end{tabular}
   \label{table:eventrate}
\end{table}


\quad 
\begin{algorithm}
\caption{Edge-Integral }
\label{alg:thesecond}
\begin{algorithmic}
	\Require $imageList$ 
	\Ensure $grayImage$
	\Function {Generator}{Image}
		\State Xtrace = [1,0,2,1,0,2,1,1,2]
		\State Ytrace = [0,2,1,0,1,2,0,1,1]
		\State imSize = size(imageList[0])
		\State H = imSize[0], W= imSize[1]
		\State SUM = zeros(H+4,W+4)
		\State T = length(ImageList)
		\For {$t = 0 \to T$}
			\State dx = Xtrace[j]
       			\State dy = Ytrace[j]
        			\State frame=imageList[t]
        			\State SUM[2-dx:2-dx+H,2-dy:2-dy+W] += frame[0]
        			\State SUM[2-dx:2-dx+H,2-dy:2-dy+W] -= frame[1]
		\EndFor
		\State $gray\_image$  = SUM
	\EndFunction	
  \end{algorithmic}
\end{algorithm}

\begin{figure*}[!t]
    \centering
    \includegraphics[width=1  \linewidth]{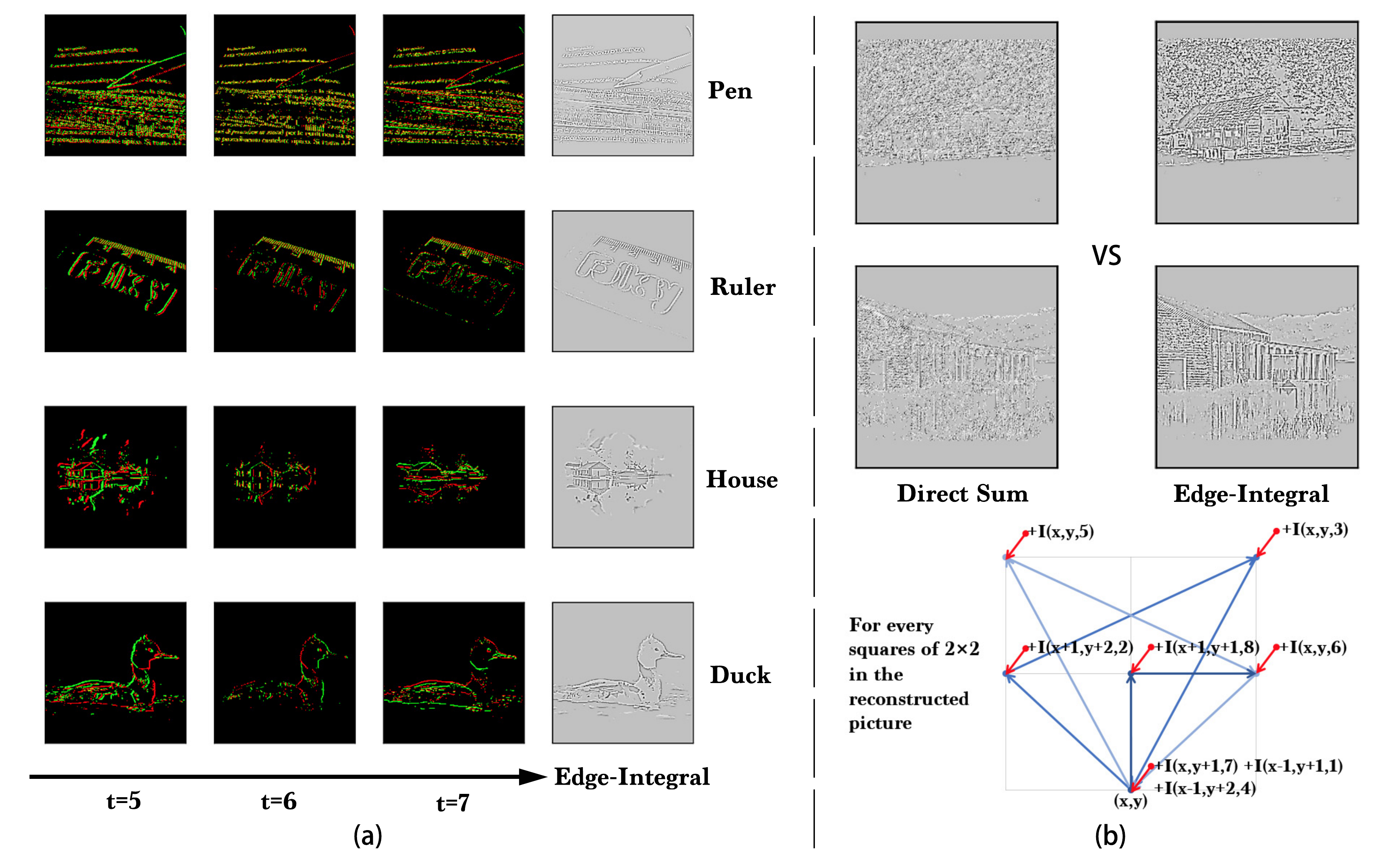}
    \caption{(a) {\bf The visualization of ES-ImageNet}. We show a few samples reconstructed from event streams and the event frames at the last three time steps for each sample. These examples are from 4 different categories and can be clearly identified. (b) {\bf Quality comparison of the reconstruction results of direct summation and Edge-Integral.} Here $I(x,y,t)$ means the intensity in (x,y) in the event frame at $time=t$.} 
    \label{fig:recovery_merge}
\end{figure*}

{\bf Visualization.} To show the quality of the data intuitively, we reconstruct the original pictures from event streams. Firstly we accumulate the events into frames, and we obtain 8 ($T=8$) event frames. Different from the traditional DVS-dataset, our dataset is generated along a fixed path with multiple directions, so when we try to reconstruct the original pictures, we need to accumulate the difference frames (so called Edge-Integral used in \cite{le1995refining}) along the opposite direction of the generating path. The results are shown in {\bf Fig. \ref{fig:recovery_merge}}, and the pseudo-code of Edge-Integral can be found in {\bf Algorithm \ref{alg:thesecond}}. More visualization DEMO can be found in supplementary files.

Analyzing the process of conversion, we know that there are 3 operations potentially causing the information loss. Firstly only the information in the V channel of HSV color space is used, and secondly, the gradient information obtained is also approximate, while at last the information is stored in low-bit. According to the method in  {\bf Fig. \ref{fig:recovery_merge}}, we are able to reconstruct the gray images, which can also be directly obtained from the original color images by the weighted sum of $(R, G, B)$. 

\begin{figure*}[!t]
    \centering
    \includegraphics[width= 0.5 \linewidth]{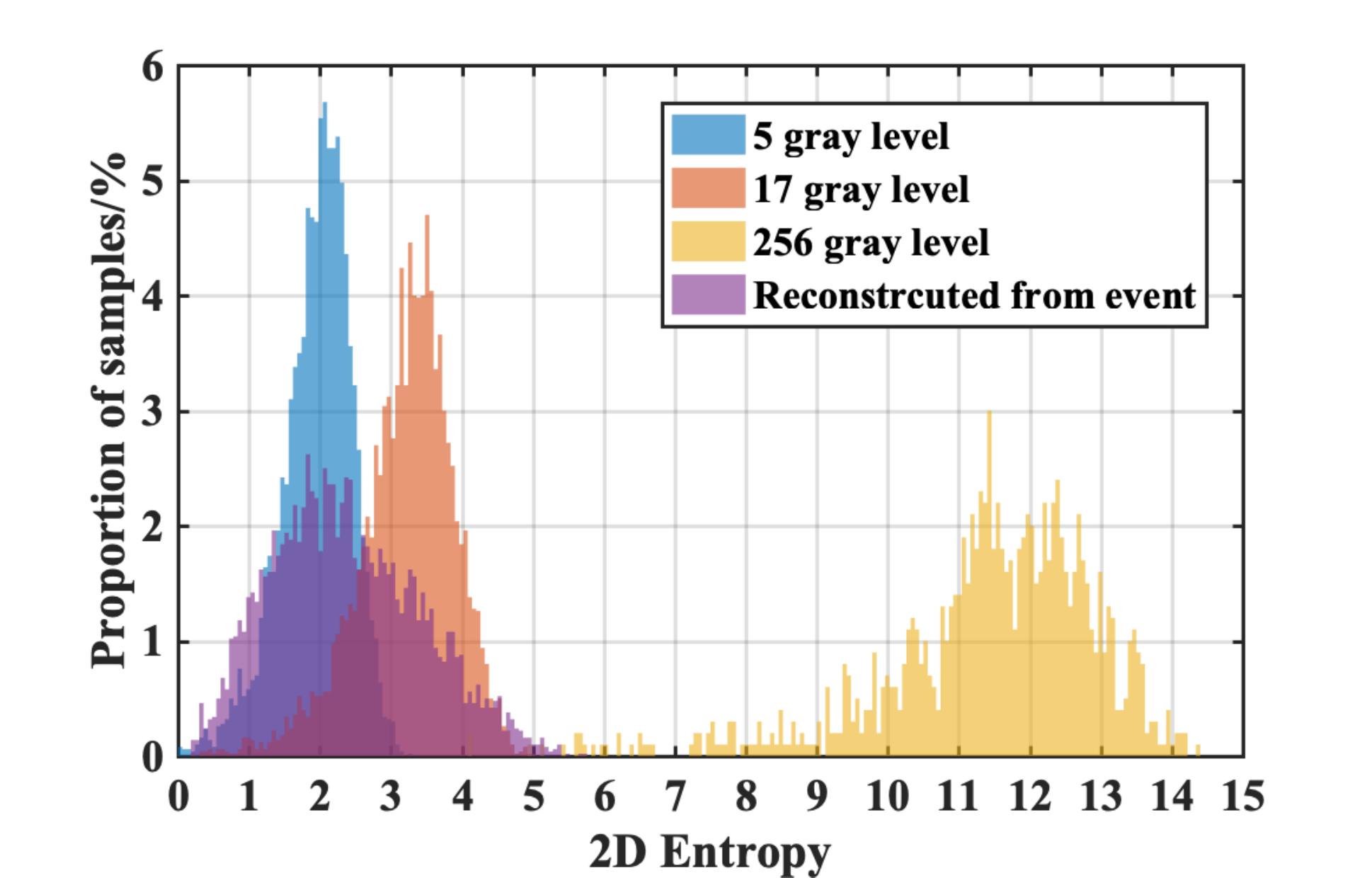}
    \caption{{\bf 2D-Entropy histogram of the different compression levels ILSVRC2012 sample groups and the reconstructed ES-ImageNet sample group}. We compare the 2D-Entropy of 4 sample groups here. The reconstructed group indicates that the samples in ES-ImageNet potentially have effective information for object classification.}
    \label{fig:Entrophy}
\end{figure*}

{\bf Information.} To further analyze the loss of information during the conversion, we still turn to the 2D-Entropy of the gray image defined in {\bf Eq. (\ref{equa:2dH})}. We randomly collect 5000  RGB-images in \emph{ILSVRC2012} (5 per class) and convert them into gray images with 256 gray level, 17 gray levels, 5 gray levels respectively. And then we find the converted samples of those RGB-images in \emph{ES-ImageNet} and reconstruct the corresponding gray images. Because the default $T=8$ in \emph{ES-ImageNet}, and each pixel value could be $0$, $1$, or $-1$, the reconstructed samples will have a total of 17 gray levels (from $0$ to $16$). 

The ordinal meaning of 2D-Entropy can tell us what level the amount of information of \emph{ES-ImageNet} is, which is no less than that with 5-gray-level compressed RGB-images and almost the same as that with 17-gray-level compressed RGB-images, as {\bf Fig. \ref{fig:Entrophy}} shows. It should be noted that the reconstruction process also causes information loss, so the original \emph{ES-ImageNet} may have more efficient information than we speculate. Considering that the application of neural morphological data does not need many high-level features, we believe that the amount of information can make this dataset a nice validation tool of SNN.


\section{Result}
\subsection*{Training Experiments}

 \begin{table}[hbt]
\centering
 \renewcommand\arraystretch{1.4}
  \centering
  \caption{Hyper-parameter setting.}
  \label{tab:exp-MNIST-pra}
\begin{tabular}{lll}
\hline& Names& Value\\ \hline
\multirow{3}{*}{Network}& 
T & 8\\
& Thresh& 0.5\\
& Decay& 0.5\\ \hline
\multirow{2}{*}{Optimizer (ADAM)}    
& Lr& 3e-2\\
& $\beta_1,\beta_2,\lambda$ & 0.9,0.999,1e-8\\ 
\hline
Activation& Lens& 0.5\\ \hline
\multirow{2}{*}{StepLR}  
& Nepoch& 10\\
& $\alpha$& 0.2\\ \hline
\multirow{2}{*}{Others} 
& BatchSize& 224$^{\rm a}$/160$^{\rm b}$\\
& Max Epoch & 50\\ \hline
\end{tabular}
\\
\footnotesize{$^{\rm a}$ Used for the training of ResNet-18.}\\
\footnotesize{$^{\rm b}$ Used for the training of ResNet-34.}
\end{table}

Because the size of this dataset is very large, it is difficult to train a classical classifier (such as K-Nearest Neighbor) on it as other DVS-dataset provides \cite{dvscifar10}. Statistical learning methods such as support vector machine (SVM) do not perform well on large-scale datasets with many categories either, and it might take days to train a vanilla nonlinear SVM on a dataset with only 500K samples \cite{svm}. To examine the quality of this dataset, we turn to four different types of deep neuron networks, two of which are ANNs while others are SNNs. The structure of ResNet-18 and ResNet-34 \cite{resnet} are applied in the experiments. The results of these experiments provide a benchmark for this dataset. It is noted that all of the accuracy mentioned here is top-1 test accuracy.

For ANNs, Two Dimension Convolutional Neural Network (2D-CNN) \cite{krizhevsky2012imagenet} has become a common tool for image classification. In order to train 2D-CNN on ES-dataset, a common approach is to accumulate the events into event frames according to the time dimension and then reconstruct the gray images \cite{wu2020liafnet} for training. Here we use Edge-Integral algorithm described in {\bf Fig. \ref{fig:recovery_merge}} for reconstruction. The network structures we use here are the same as those in the original paper \cite{resnet}.

Meanwhile, regarding the time dimension as the depth, this dataset can also be considered as a video dataset, so the classic video classification methods can also be utilized, like 3D-CNN \cite{3DCNN,hara2018can}. By introducing convolution of depth dimension, 3D-CNN has the ability of processing time-domain information. The structures we used are 3D-ResNet-18 and 3D-ResNet-34, and convolution kernel is chosen to be $3 \times 3 \times 3$, which ensures that the largest receptive field of the network can cover the whole time (depth) dimension.

\begin{figure*}[!t]
    \centering
    \includegraphics[width= 0.8  \linewidth]{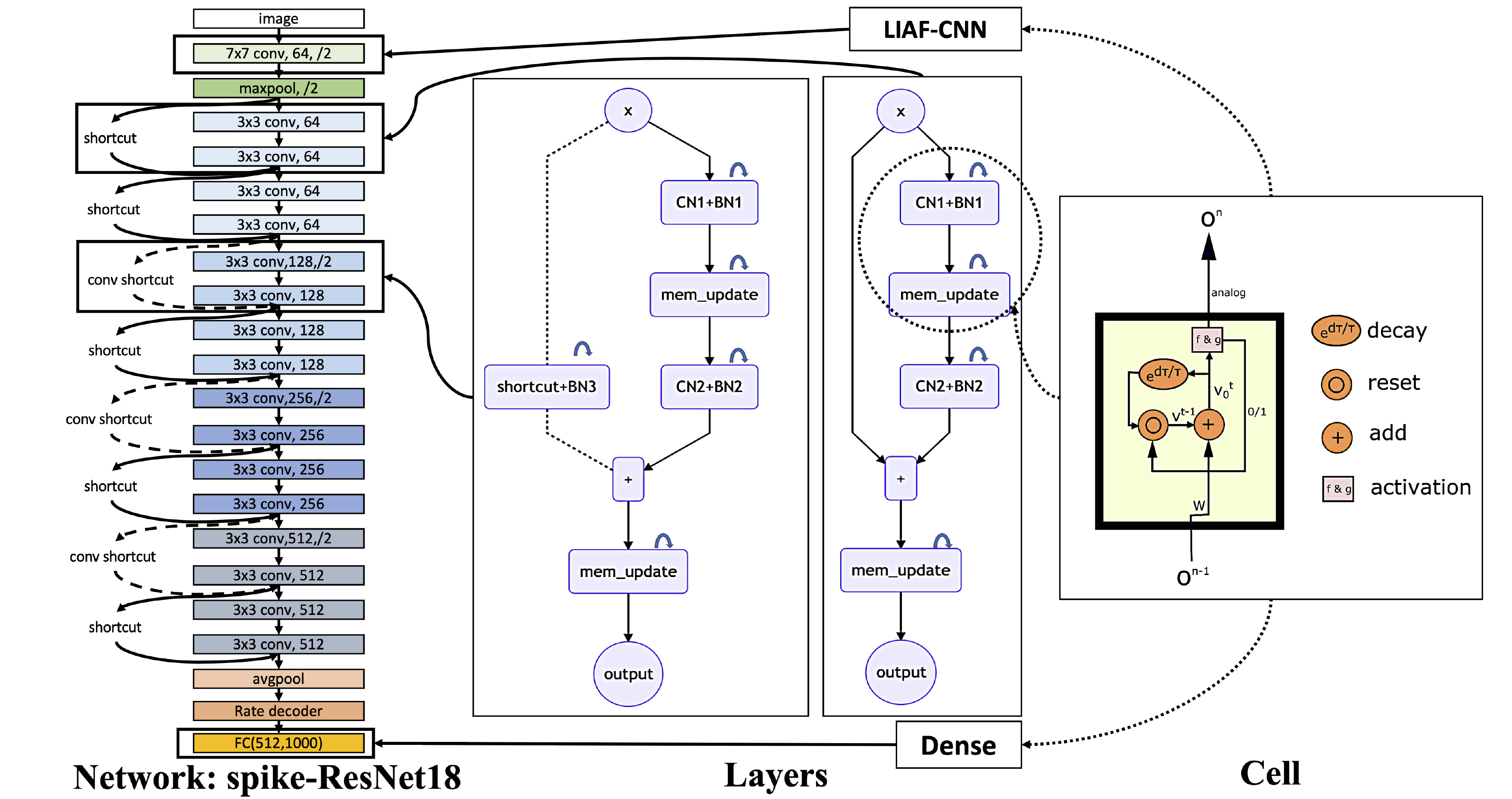}
    \caption{{\bf SNN structure used in the experiments.} On the right, we show the internal structure of LIAF neurons. By changing synaptic connections, we can obtain a variety of layer structures, where CN denotes the Convolutional layer and BN denotes the 3D-BatchNorm layer. Using these layers, we can build a scalable LIAF residual network structure.}
    \label{fig:LIAF-ResNet}
\end{figure*}

For SNNs, we choose an SNN based on Leaky Integrate-and-Fire (LIF) neurons \cite{LIF} and an SNN based on Leaky Integrate-and-Analog-Fire (LIAF) \cite{LIAF} neurons. Rate coding  \cite{ratecoding} is used to  decode the event information because the significance of the specific time when the spikes appear in this dataset is weaker than the number of spikes.  Both of the SNN models are trained using STBP method  \cite{STBP} and sync-Batch Normalization  \cite{ioffe2015batch}, and the network structures similar to ResNet-18 and ResNet-34 are built as shown in {\bf Fig. \ref{fig:LIAF-ResNet}}. The basic LIF \cite{LIF} model is described in {\bf Eq. (\ref{equa:BLIF})},

\begin{equation} 
\label{equa:BLIF}
U=\left\{
\begin{aligned}
&\tau_m\frac{dU}{dt}=-U+E_L+R_mI_e & ~ & if~~U<U_{thresh}\\
&U_{reset} & ~ & U \geq U_{thresh}\\
\end{aligned}
\right.
\end{equation}

where U is the membrane potential, $E_L$ is adjusted to make the resting potential match that of the cell $L$ being modeled. $I_e$ is the input current and the $R_m$ is membrane resistance. $U_{reset}$ is a parameter adjusted according to the experiment data, and $\tau_m$ is the membrane time coefficient. The LIF neuron will fire a spike when $U$ reaches the $U_{thresh}$, and the spike can be \{0,1\} in LIF or an analog value in LIAF. Solving the model, we have the $U(t)$ shown in {\bf Eq. (\ref{equa:U(t)})}.
 
\begin{eqnarray}
\label{equa:U(t)}
U(t)= E_L+R_mI_e+(U(0)-E_L-R_mI_e)e^{-t/\tau_m}.
\end{eqnarray}

This equation doesn't take the reset action into consideration. For large scale computer simulation, simplification is needed on this model and using the discrete LIAF/LIF model. Using $l$ to present the layer index and $t$ for the time, the LIAF model can be described by the following equations

 \begin{eqnarray}
 \label{LIAF-diedai}
 & u_0^{t,l}=u^{t-1,l}+h(o^{t,l-1})+b^l\\
 \label{LIAF-diedai1}
 & s^{t,l} = f(u_0^{t,l})\\
\label{LIAF-diedaik}
 & o^{t,l} = g(u_0^{t,l})\\
 & u^{t,l}= u_0^{t_k,l}d(s^{t,l})
 \label{LIAF-diedai2}
 \end{eqnarray}
 
where $h$ is the weighted sum function of the input vector $o^{t,l-1} $, which is related to the specific connection mode of synapses and is equivalent to $R_mI_e$. $s$ is the spike used to reset the membrane potential to $U_{reset}$ and $o^{t,l}$ is the output of neurons to the next layer. We often use $d(x) = \tau (1-x) $ for simplification in this model, where $d(x)$ describes the leaky processing and $\tau$ is a constant relative to $\tau_m$. $f$ is usually a threshold-related spike function, while $g$ is selected to be a commonly used continuous activation function. If $g$ is chosen as the same function as $f$, then the above model is simplified to the LIF model as

 \begin{eqnarray}
 \label{LIF-diedai}
 & u_0^{t,l}=u^{t-1,l}+h(o^{t,l-1})+b^l\\
  \label{LIF-diedai1}
 & o^{t,l} =s^{t,l} = f(u_0^{t,l})\\
 & u^{t,l}= u_0^{t_k,l}\tau (1-s^{t,l})
 \label{LIF-diedai2}
 \end{eqnarray}
 
To build a Spiking-ResNet model, we proposed the spiking convolutional layer and spiking-ResBlock structure. Only the $h$ in {\bf Eq. (\ref{LIAF-diedai})} and {\bf Eq. (\ref{LIF-diedai})} needs to be changed to become different types of SNN layers. For full-connection layer (or Dense), we choose $h(o^{t,l-1})=W_{l}*o^{t,l-1}$, where $W_{l}$ is the weight matrix of $l$ layer. In convolutional layer, $h(o^{t,l-1})=W_{l}\otimes o^{t,l-1}$, where $\otimes$ is convolution operation. 

The residual block structure we used in SNN is a little bit different. For a better performance in deep SNN training, we add a 3D-BatchNorm layer on membrane potential, where we treat the temporal dimension in SNN as the depth of the general 3D-data. In {\bf Fig. \ref{fig:LIAF-ResNet}}, CN denotes convolutional layer and BN denotes 3D-BatchNorm layer, the \emph{mem\_update} layers are described by {\bf Eq. (\ref{LIAF-diedai1})-(\ref{LIAF-diedai2})} in LIAF-ResNet , and {\bf Eq. (\ref{LIF-diedai1})-(\ref{LIF-diedai2})} in LIF-ResNet. To  keep the coding consistency, before each output of residual block, we add a \emph{mem\_update} layer.

The best test results are obtained based on the same set of hyper-parameters and different random seeds, which are shown in the {\bf Table \ref{tab:exp-MNIST-pra}}, and the results are listed in the {\bf Table \ref{tab:result}}. During the training, the initial learning rate is $0.03$, the optimizer is ADAM  \cite{adam}, and the learning rate is optimized by StepLR learning schedule. NVIDIA-RTX2080Tis are used for training and Pytorch  \cite{Pytorch} deep-learning framework is used for programming for all of those experiments. 

\quad
\subsection*{Test Results}

\begin{figure*}[!t]
    \centering
    \includegraphics[width=0.6 \linewidth]{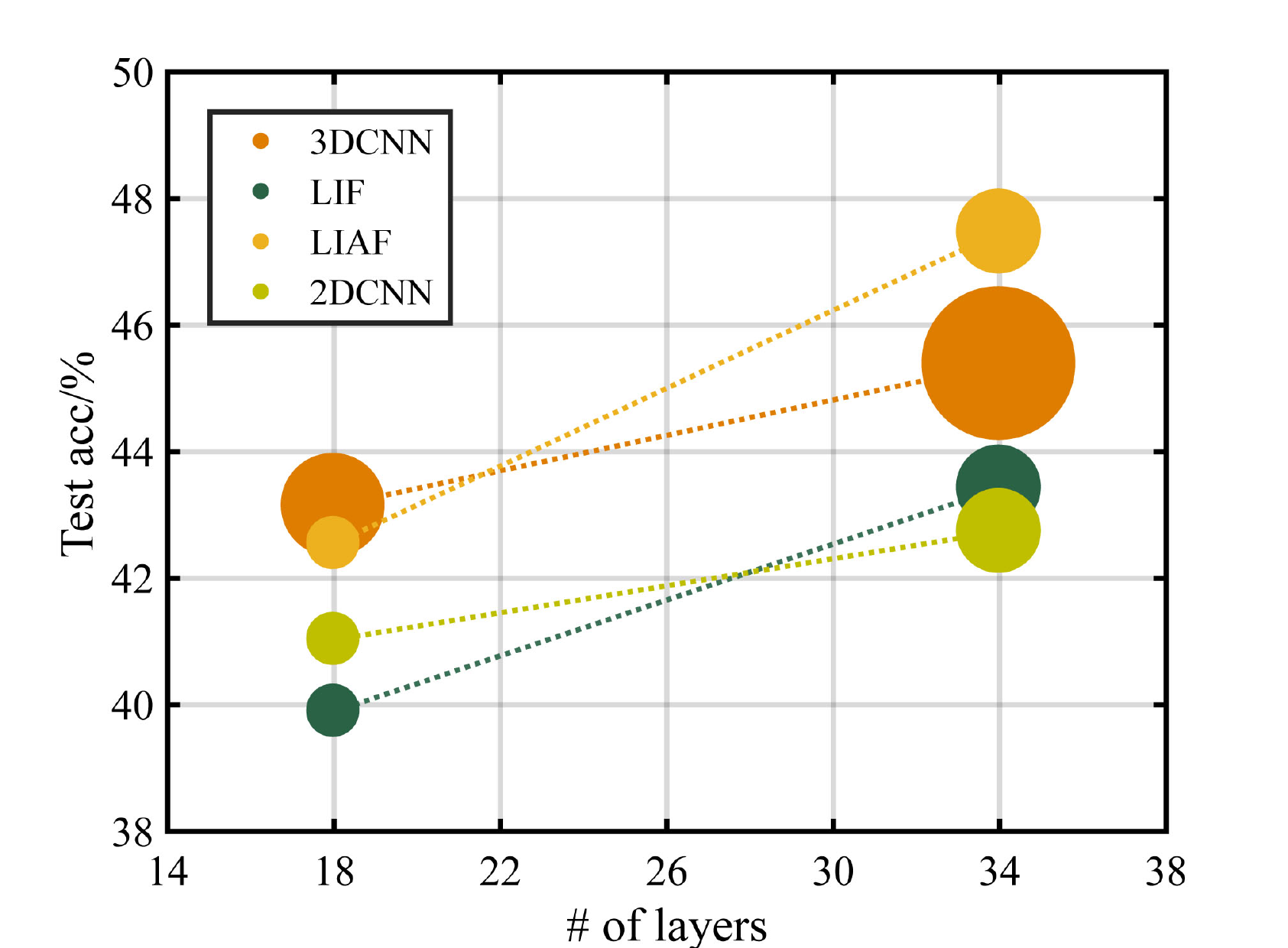}
    \caption{{\bf Testing accuracy with the structure of ResNet-18 and ResNet-34}. The radius of the data points represents the relative size of the parameters of the networks}
    \label{fig:result}
\end{figure*}

 \begin{table}[hbt]
 \centering
  \centering
   \renewcommand\arraystretch{1.4}
 \caption{Test results \& Benchmarks.}
 \label{tab:result}
\begin{tabular}{|c|c|c|c|}
\hline
\multicolumn{1}{|l|}{\textbf{Structure}} & \multicolumn{1}{c|}{\textbf{Type}} & \multicolumn{1}{l|}{\textbf{Test Acc/\%}} & \multicolumn{1}{l|}{\textbf{ \# of Para}}\\ \hline
\multirow{4}{*}{ResNet18}  & 2D-CNN & 41.030 & 11.68M \\ \cline{2-4} 
                                      & 3D-CNN &\textbf{43.140}&28.56M\\ \cline{2-4} 
                                      & LIF (baseline) & 39.894 & 11.69M \\ \cline{2-4} 
                                      & LIAF & 42.544 & 11.69M  \\ \hline
\multirow{4}{*}{ResNet34}& 2D-CNN & 42.736 & 21.79M \\ \cline{2-4} 
                                     & 3D-CNN & 45.380 & 48.22M \\ \cline{2-4} 
                                     & LIF (baseline) & 43.424 & 21.80M \\ \cline{2-4} 
                                     & LIAF & \textbf{47.466} & 21.80M \\ \hline
\end{tabular}
\end{table}

As {\bf Table \ref{tab:result}} shows, the highest test accuracy based on ResNet-18 structure is obtained by 3D-CNN, which is $43.140\%$. And the best result on ResNet-34 reaches $47.466\%$ obtained by LIAF-SNN. The experimental results of LIF-SNN, which is the traditional SNN model, will provide a baseline for this dataset, and we expect more advanced and large-scale SNNs or other neuromorphic algorithms to be tested on this dataset. 

\section{Discussion}

\subsection*{Performance}

Observing the results, we will find that the SNN models can obtain a relatively high classification accuracy with fewer parameters. The sparsity of the data in \emph{ES-ImageNet} may lead to this phenomenon, for SNN can deal with spatiotemporal information efficiently, and a large number of parameters in ANNs-based video classification algorithm (like 3D-CNN tested in this article) may cause over-fitting on this dataset.

We find the other two reasons for the accuracy loss. The wrongly labeled samples may also seriously interfere with the training progress. This problem is obvious in \emph{ImageNet}  \cite{northcutt2019confidentlearning}, and we also found this problem when we conduct a manual inspection, but there is no good method for efficient and accurate screening now. Another problem is the information loss. Given this problem, we propose several possible ways to optimize it. One is to filter out more samples with the highest and lowest information entropy (representing the largest noise rate and the smallest amount of information respectively) in the training set. The other is to increase the number of time steps of the transformation, but it will increase the storage cost.

It should be noted that in the experiments we do not use any data augmentation method. In fact, placing event frames in random order, using dynamic time length, or process each frame with random clipping are acceptable on this dataset and may bring significant performance boost. Research is under way on such data augmentation and pre-training technologies, and we hope more related research can use this dataset.

\subsection*{Computation Cost}

To make a more objective comparison, we also count and compare the theoretically minimum number of FP32 operands required by the feed-forward process of these networks with measuring the power consumption in a field-programmable gate array (FPGA).

Here we compare the number of necessary calculation operands required by the feed-forward process of eight different networks used in the main article. It should be noted that we calculate the number of floating-point multiplication operands and floating-point addition operands separately (not MACs) and the operands of normalization layers are not included in calculation. 

{\bf 2D-CNNs} use the ResNet structures with 18/34 layers, and most of the operands are contributed by convolution layers. In this work, we compress and reconstruct the 4-dimensional event data in \emph{ES-ImageNet} into 2-dimensional gray images, then feed them into 2D-CNNs. The process is then the same as the way we train a ResNet on \emph{ImageNet}. In the network, the dimensions of the features change in the following order: $[1(channel)\times224(width)\times224(height)] \to(maxpooling)[64\times110\times110]\to[64\times55\times55]\to[128\times28\times28]\to[256\times14\times14]\to[512\times7\times7]\to[512]\to[1000]$

{\bf 3D-CNNs} considers the depth dimension \cite{3DCNN}, and treat this dataset as a video dataset \cite{3DCNN-video}, so the feature is kept in 4-dimension in ResBlocks. In the network, the dimensions of the features change in the following order: $[2(channel)\times8(depth)\times224(width)\times224(height)] \to(maxpooling)[64\times8\times110\times110]\to[64\times4\times55\times55]\to[128\times2\times28\times28]\to[256\times1\times14\times14]\to[512\times1\times7\times7]\to[512]\to[1000]$

The training procedure of {\bf LIF-SNNs} is like running a 2D-CNNs eight  along with the processing of the last moment of membrane potential memory information and the spikes inputs for every layer, then average the spike trains in time dimension in the final linear layer to decode the spiking rate. These networks keep the data in 4-dimension with $T=8$ unchanged until the decoding layer, so the dimensions of the features change in the following order: $[2(channel)\times8(T)\times224(width)\times224(height)]\to(max~pooling)[64\times8\times110\times110]\to[64\times8\times55\times55]\to[128\times8\times28\times28]\to[256\times8\times14\times14]\to[512\times8\times7\times7]\to[8(depth)\times512](rate~decoded)\to[512]\to[1000]$. 

The training procedure of {\bf LIAF-SNNs} is quite the same as {\bf LIF-SNNs}, the only different with {\bf LIF-SNNs} is that it doesn't use binary spikes to convey information between layer, however, it use analog spike. The dimensions of the features are the same as the ones in {\bf LIF-SNNs}.

It is worth noting that since the input of {\bf LIF-SNNs} is only 0 and 1, convolution does not need to compute floating-point multiplication, but addition under limited combination. As a large number of zeros appear in the input of each layer of {\bf LIF-SNNs}, the optimization of sparse input for {\bf LIF-SNNs} has become a formula in SNN accelerators. Therefore, in order to make a fair comparison, we can use the average fire rate obtained in the experiment multiplied by the input of SNN as the proportion of the number of floating-point numbers that need to participate in the addition calculation(FP32 +), so as to estimate the actual amount of computation of SNNs and CNNs. An observation is that the fire rate always shows a downward trend with the increase of training epochs, which means a decrease of meaningless spikes.

 \begin{figure*}[hbt]
    \centering
    \includegraphics[width= 0.6  \linewidth]{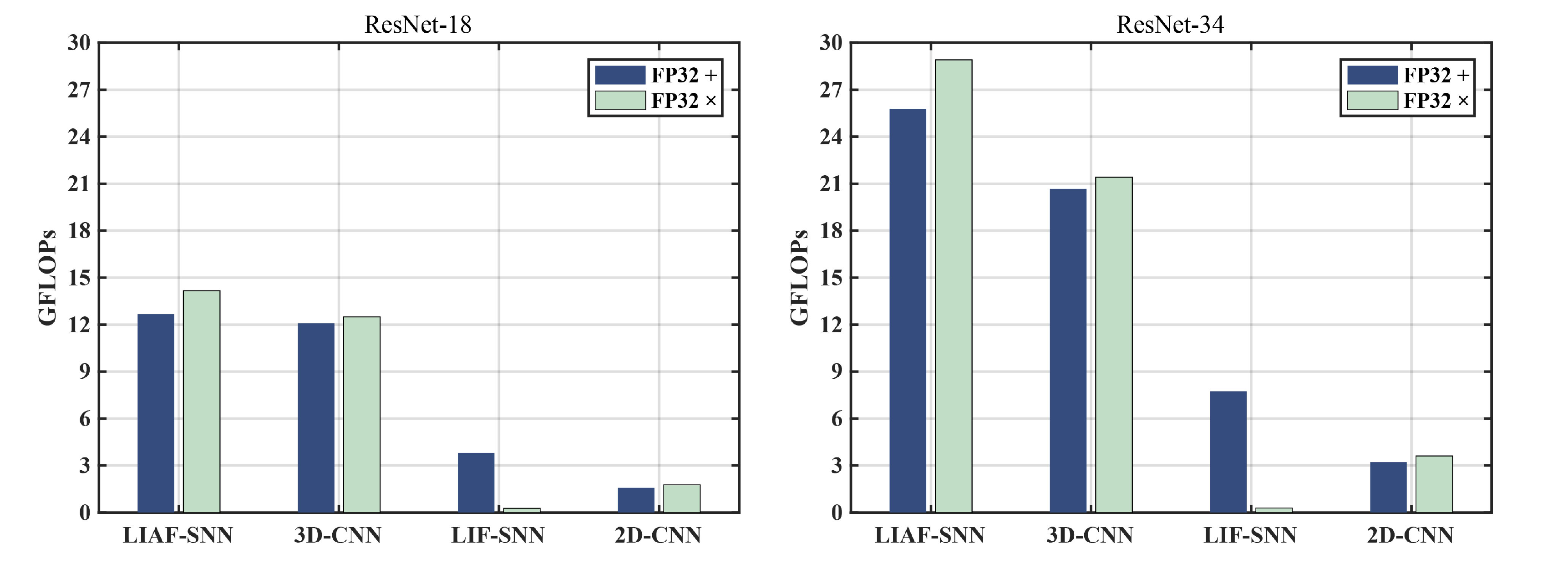}
    \caption{{\bf The comparasion of FP32 addition (FP32 +) and FP32 multiplication (FP32 x) operations in feedforward process between the models we use in the experiments}. It should be noted that the number of FP32+ operands of LIF have been multiplied by a sparsity factor (30\%) for a fair comparison.}
    \label{fig:ops}
\end{figure*}

In this experiment, the initial fire rate of SNN is no larger than 30\%, and with the increase of training epochs, the fire rate would gradually reduce to less than 10\%. {\bf To compare the results of SNN in the worst case, we take 30\% as the sparsity rate.} Base on these conditions we can get {\bf Fig. \ref{fig:ops}}.

 \begin{figure*}[hbt]
    \centering
    \includegraphics[width= 0.6   \linewidth]{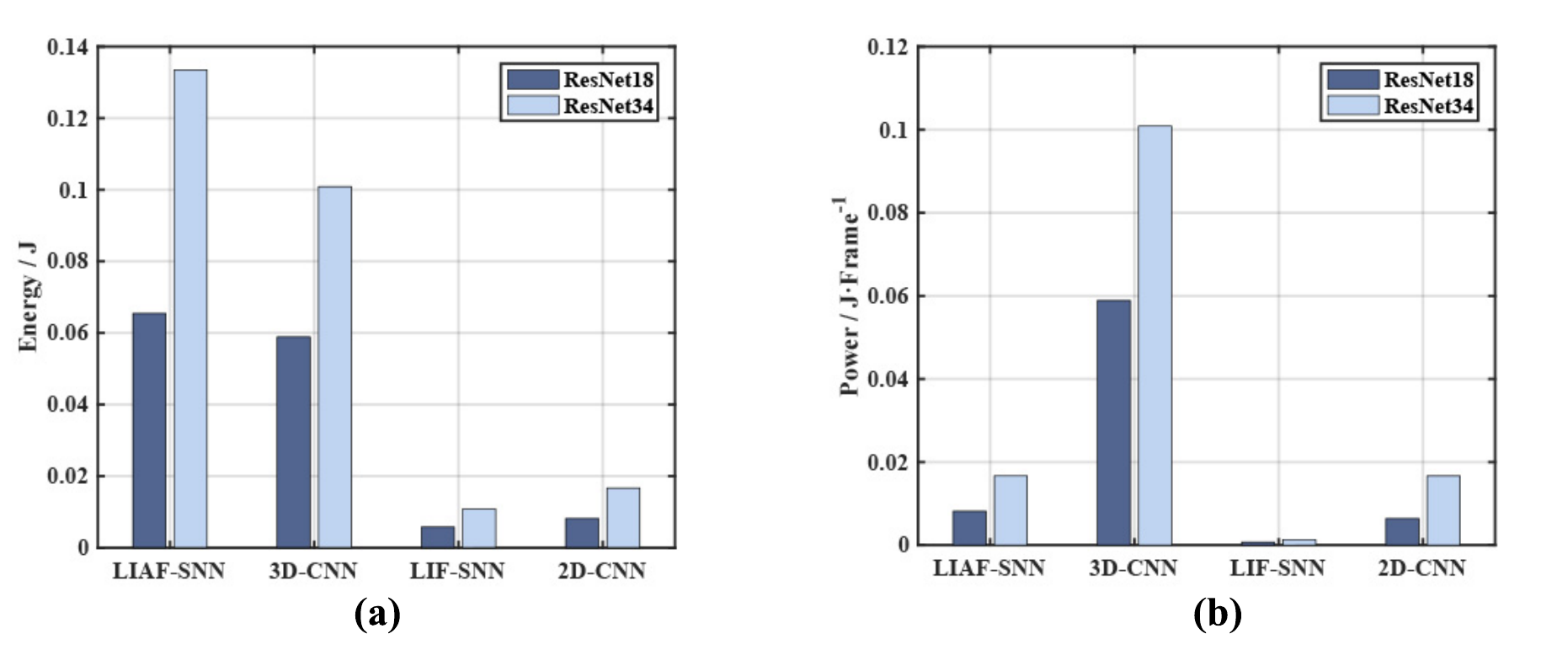}
    \caption{{\bf Energy and power consumption in the experiments}. (a) The comparison on energy cost. The LIF-SNNs have shown a significant advantage in energy consumption, whose energy cost is the half of that of CNN. (b) The power of each model in the \emph{Frame} time unit, which is also the energy required for one feed-forward process.}
    \label{fig:energy}
\end{figure*}

One of the advantages of SNN comparing with ANN is its power consumption  (especially in SNNs' accelerators). On FPGAs the SNNs could have a significant power advantage if the training algorithm is well designed. The data in \cite{L1norm} about the basic operands' power consumption can provide an estimation of the power consumption of the networks in the experiments. Each FP32 + operation requires  1.273 pJ energy, and each FP32 × operation requires  3.483 pJ energy. Then we can get the result in {\bf Fig. \ref{fig:energy} (a)}. In addition, we also give the power comparison commonly used for SNNs \cite{deng2020rethinking} in {\bf Fig. \ref{fig:energy} (b)}, where we calculate the energy for each feed-forward process (so we call it power). It should be noted that SNNs need $T$ frames to give one prediction, and both 3D-CNN and 2D-CNN give one prediction based on one frame.

These results also support the SNNs' energy advantages in this task. For these reasons, we think \emph{ES-ImageNet} would be an SNNs friendly dataset. We still hope that more ANNs algorithms will be proposed to solve these challenges elegantly and efficiently, which may also provide guidance for the development of SNNs.

\subsection*{Limitation}

For the conversion algorithm, we generate temporal features by applying artificial motion to static images like most conversion methods, which is still different from the real scene. It is the limitation for those dynamic datasets derived from static data. In addition, in order to compress the volume of the dataset and extract more information, we reduce the randomness of data during generation, thus losing a certain feature of DVS cameras recording but being more friendly to SNNs.

In the analysis part, due to the limitation of mathematical tools, the 2D-Entropy we adopt can reflect only the amount of information, not the amount of effective information. Therefore, it can only be used as a reference rather than a standard. In addition, the reconstruction method and the compression method used in measurement would influent the information, though we have compared them as fairly as possible.

In the training method, due to the limitation of hardware conditions and algorithms, we can only provide the benchmarks of SNNs and ANNs based on ResNet-18 and ResNet-34. It is hoped that more research will participate in the training of larger and better models.

\section{Conclusion}

In this paper, we provide a new neuromorphic vision dataset named \textit{ ES-ImageNet} for event-based image classification and validation of SNN algorithms. We proposed a method called ODG, transforming a famous image classification dataset \emph{ILSVRC2012} into its event-based version with a method called Edge-Integral to reconstruct the corresponding gray images based on these event streams. ODG method includes a carefully designed image movement, which results in the value changes in the HSV color space and provides spatial gradient information. This algorithm can efficiently extract the spatial features to generate event streams. 

For testing the properties of datasets, we use the Edge-Integral method to exhibit some of the reconstructed samples, and also calculate the 2D-Entropy distribution of the dataset. Furthermore, a comparative experiment is conducted using 2D-CNN, 3D-CNN, LIF-SNN, and LIAF-SNN, and these results provide a  benchmark for later research, also support that this dataset is a suitable validation tool for SNNs.

This dataset solves the problem of lacking a suitable large-scale classification dataset in the SNNs' research field. It fills in the gap of lacking a suitable dataset for the verification of large-scale SNNs so that the corresponding algorithm is expected to be better optimized, more SNNs' structures and training algorithms will be explored, thereby promoting practical applications of SNNs.

\section*{Conflict of Interest Statement}

The authors declare that the research was conducted in the absence of any commercial or financial relationships that could be construed as a potential conflict of interest.

\section*{Author Contributions}

\textbf{Author contributions:}
Y. Lin and G. Li conceptualized the work.
Y. Lin designed research.
Y. Lin, W. Ding and S. Qiang designed and conducted the experiment.
Y. Lin  analyzed data.
L. Deng, G. Li supervised the work.
All authors wrote the manuscript.
\bigskip

\section*{Funding}
This work is partially supported by National Key R\&D Program of China (2018AAA0102604, 2018YFE0200200),  and  National Science Foundation of China (61876215), and Beijing Academy of Artificial Intelligence (BAAI), and a grant from the Institute for Guo Qiang, Tsinghua University, and Beijing Science and Technology Program (Z191100007519009), and the open project of Zhejiang laboratory, and the Science and Technology Major Project of Guangzhou (202007030006).

\section*{Supplemental Data}
A short movie for data visualization is provided as \emph{DVSwithRGB.avi}. \\
The source codes are also available for using this dataset.\\

\section*{Data Availability Statement}
The datasets ES-ImageNet (100GB) for this study can be download in the \url{https://cloud.tsinghua.edu.cn/d/94873ab4ec2a4eb497b3/}.\\
The converted event-frame version (40GB) can be found in \url{https://cloud.tsinghua.edu.cn/d/ee07f304fb3a498d9f0f/}.\\
The codes can be found in \url{https://github.com/lyh983012/ES-imagenet-master}.\\




\end{document}